\documentclass{article}

\usepackage[final]{neurips_2024}
 \setcitestyle{numbers}

\usepackage{todonotes}
\usepackage{wrapfig}

\usepackage[utf8]{inputenc} %
\usepackage[T1]{fontenc}    %
\usepackage{hyperref}       %
\usepackage{url}            %
\usepackage{booktabs}       %
\usepackage{amsfonts}       %
\usepackage{nicefrac}       %
\usepackage{microtype}      %
\usepackage{xcolor}         %

\usepackage{graphicx} %
\usepackage{subcaption}
\usepackage{booktabs}

\title{Neural Normalized Compression Distance and the Disconnect Between Compression and Classification }

\author{%
  John Hurwitz \\
  University of Maryland, Baltimore County\\
  Laboratory for Advanced Cybersecurity Research\\
  \And
  Charles Nicholas \\
  University of Maryland, Baltimore County\\
  \And
  Edward Raff\\
  Booz Allen Hamilton\\
  University of Maryland, Baltimore County\\
}

\begin{document}

\maketitle

\begin{abstract}
It is generally well understood that predictive classification and compression are intrinsically related concepts in information theory. Indeed, many deep learning methods are explained as learning a kind of compression, and that better compression leads to better performance. We interrogate this hypothesis via the Normalized Compression Distance (NCD), which explicitly relies on compression as the means of measuring similarity between sequences and thus enables nearest-neighbor classification. By turning popular large language models (LLMs) into lossless compressors, we develop a \textit{Neural} NCD and compare LLMs to classic general-purpose algorithms like gzip. In doing so, we find that classification accuracy is not predictable by compression rate alone, among other empirical aberrations not predicted by current understanding. Our results imply that our intuition on what it means for a neural network to ``compress'' and what is needed for effective classification are not yet well understood.
\end{abstract}

\section{Introduction} \label{introduction}
The link between compression and prediction has been well-established \cite{mackay2003information}. Predictors (such as machine learning models) can be turned into compressors when combined with an entropy coding technique such as arithmetic coding \cite{arithmetic_coding}. Conversely, compressors can be used to make predictions via a compressor-based distance metric and a distance-based learning algorithm such as $k$-nearest neighbors. A popular method for the latter approach is the Normalized Compression Distance (NCD)~\cite{li2004similarity}, which is theoretically optimal as a distance metric if given an optimal compressor. NCD has seen success in a variety of domains, including anomaly detection and clustering \cite{Keogh2004}, text classification \cite{jiang-etal-2023-low}, and image classification \cite{jiang_few-shot_2022}. Here, the compressors used are primarily traditional compression algorithms such as \textit{gzip} and \textit{lzma}. While these traditional compression algorithms can compress data quickly and successfully in an impressively wide variety of domains, much better lossless compression performance is generally possible in any given specialized domain through learned neural network-based compressors \cite{bellard2021nncp,cmix,tensorflow-compress}. In the text domain, the leading compressors in terms of compression rate are neural-network based \cite{mahoney_large_text_compression} and far surpass the performance of traditional compressors, at the cost of being more computationally expensive. 

Given the superior compression ability of neural compressors, the natural question arises of whether their improved compression rates translate into improved predictive performance. Indeed, it has been hypothesized that compression algorithms with better compression rates should classify better on account of closer approximation to Kolmogorov complexity. We investigate  NCD-based classification performance differences by replacing traditional compressors with LLM-based neural compressors in this paradigm, noting the effect of compression rate on test accuracy, and comparing NCD performance with Euclidean distance with a model's latent representation. We find that counter to conventional wisdom, NCD-based classification accuracy is not predictable solely from compression rate. We exhibit cases where a neural compressor achieves consistent and superior compression rates across datasets but can either outperform or underperform traditional compressors depending on the dataset. 

This paper is organized as follows. In \autoref{method} we describe our novel approach to doing NCD-based text classification via pretrained LLMs as neural compressors in the few-shot text classification setting, and provide requisite background in the information-theoretic grounding of compression distance. In \autoref{related} we highlight related work in neural compression, compression-distance-based machine learning, and the connection between compression rate and classification accuracy. In \autoref{experiments} we describe our experimental results and three primary findings relating to the relationship between compression rate and accuracy, effects of varying the choice of LLM for neural compression, and a comparison to using the models' latent representations with Euclidean distance. We then provide a brief conclusion in \autoref{conclusion}.

\section{Method: Neural Normalized Compression Distance} \label{method}

In this section we provide a brief description of Kolmogorov complexity and its connection to practical notions of compression distance, arithmetic coding and its usefulness in lossless compression schemes, and our novel approach to using pretrained LLMs as neural compressors in the few shot text classification setting. 

\subsection{Kolmogorov Complexity and Compression Distance}
Given a sequence $x$, the Kolmogorov Complexity $K(x)$ \cite{kolmogorov1963} is defined as the length of the shortest computer program that outputs $x$. $K(x)$ can be interpreted as the output of the best possible lossless compressor given any input bitstring $x$. The Conditional Kolmogorov Complexity function $K(x|y)$ is defined as the length of the shortest program that outputs $x$ given another sequence $y$ as input. Using the notion of Kolmogorov Complexity, Li et al. \cite{li2004similarity} define the Normalized Information Distance (NID).

\begin{equation}
\textnormal{NID}(x, y) = \frac{\max \{ K(x|y), K(y|x) \}}{\max \{ K(x), K(y) \}}
\end{equation}

The NID is a metric\footnote{The proof that NID is a metric includes negligible error terms for the identity axiom and triangle inequality \cite{li2004similarity}.} and has a range of $[0,1]$. The lower the NID, the more shared information exists between $x$ and $y$ which gives an indication of similarity. It is well-known that $K$ is uncomputable. In practice, we can approximate $K$ by using any compression algorithm. Li et al. \cite{li2004similarity} define the Normalized Compression Distance such that, given a compression algorithm, and a function $C(x)$ which returns the length of the compressed output of sequence $x$ in bytes, gives us a practical approximation of the NID. $xy$ indicates the concatenation of sequences $x$ and $y$. 

\begin{equation}
\textnormal{NCD}(x, y) = \frac{C(xy) - \min \{ C(x), C(y)\}}{\max \{ C(x), C(y)\}}
\end{equation}

\subsection{NCD-Based Sequence Classification}
By using any compressor to calculate pairwise NCDs, these NCDs can be used as the distance metric in the non-parametric classification algorithm $k$ nearest neighbors ($k$NN). There is an $O(n^2)$ complexity for the pairwise distance calculations between the train set and test set, each requiring the compression of a concatenated pair of samples. In current practice, this complexity is a limiting factor for choice of model size and dataset size. To use a compressor in this way only requires the compression phase; decompression is never needed.

\subsection{Arithmetic Coding}
In modern neural lossless compression paradigms, arithmetic coding \cite{arithmetic_coding} is one of the most popular schemes for entropy coding due to its near-optimality, and its ability to handle adaptive probabilistic models \cite{valmeekam2023llmziplosslesstextcompression, delétang2024languagemodelingcompression, bellard_ts_zip, cmix, bellard2021nncp}. Arithmetic coding is an entropy coding technique which, given a probability distribution as input, represents a sequence of symbols as a single arbitrary-precision floating-point number between 0 and 1. Symbols with high probability will be encoded with fewer bits than symbols with low probability. Arithmetic coding is compatible with an adaptive probabilistic model where the next-symbol probabilities can differ for each index in the sequence. Language models fit precisely this paradigm, where the prediction at each index of a sequence yields a probability distribution for the next symbol. By doing arithmetic coding on this probability distribution as input, we perform lossless compression. The resulting bitstream, with access to the same probabilistic model, can be used in a reverse process to reconstruct the original sequence.

\subsection{Neural Normalized Compression Distance for Sequence Classification}
We refer to the use of neural network-based compressors to calculate NCDs as Neural Normalized Compression Distance (Neural NCD). We extend previous work on compression-based sequence classification by using Neural NCD in the few-shot text classification setting. The neural compressor, which is a language model paired with arithmetic coding, is used to calculate NCDs that are then used as the distance metric for $k$NN. There are two primary reasons why using pretrained LLMs as the probabilistic models in a lossless compression scheme is useful for studying the relationship between compression rate and accuracy of NCD-based classification. First, neural compressors can achieve much better compression rates than traditional compressors in specific domains, allowing us to investigate the effects of compression rate on classification accuracy when using neural compressors. Second, the model can be easily swapped out, allowing us to study similar effects when varying model architecture and size.

\section{Related Work} \label{related}
{\textbf{Neural Compression:}} Neural networks have been successfully used to perform lossless compression on a variety of domains including text \cite{bellard_ts_zip, bellard2021nncp, cmix} and images \cite{jiang_few-shot_2022}. Large Language Models (LLMs) in particular have shown promise for lossless compression of text \cite{valmeekam2023llmziplosslesstextcompression, delétang2024languagemodelingcompression} due to excellence in next-symbol prediction, with neural methods leading the Large Text Classification Benchmark \cite{mahoney_large_text_compression} and achieving better compression rates than traditional compressors like \textit{gzip}, at the cost of more computation. Other works that attempt alternative architectures inspired by compression algorithms are beyond the current scope of this article~\cite{pmlr-v187-saul23a}.

{\textbf{Compression Distance for Machine Learning:}} There is empirical evidence to show that compression distance is effective in anomaly detection, clustering, and classification of sequential data \cite{Keogh2004, cebrian2005common}. Li et al. \cite{li2004similarity}, who proposed NID, also showed its effectiveness in text classification. Jiang et al. \cite{jiang-etal-2023-low} show that NCD-based classification with the traditional compressor \textit{gzip} performs competitively with neural methods such as BERT \cite{devlin-etal-2019-bert} on few-shot text classification tasks. A follow-up work \cite{jiang_few-shot_2022} uses deep latent variable models for compression of images, outperforming supervised neural network methods for few-shot image classification. Others have relaxed the compression requirements and instead extracted feature-vectors from compression dictionaries/methods to obtain faster distance calculations~\cite{raff_lzjd_2017,Raff2020}. 

{\textbf{Compression Rate and Classification Accuracy:}} Jiang et al. \cite{jiang-etal-2023-low} show a moderate linear correlation between compression rate and test accuracy among traditional compressors for few shot text classification, while noting the importance of the actual compression algorithm and finding cases where algorithms with better compression rates still under-perform ones with lower ratios. A follow-up work \cite{jiang_few-shot_2022}  also shows a similar correlation among mostly traditional compressors and one latent variable neural compressor on image classification datasets, and older work in malware analysis found a similar result\cite{Borbely2015}. Notably, compression similarity has seen wide use in malware for its ability to handle hard-to-parse and large binary files~\cite{10.5555/1370628.1370630,SResende2019,Raff2020,raff_lzjd_digest,raff_lzjd_2017,raff_shwel}.

\section{Experiments} \label{experiments}

Three key insights are derived from our study. First, accuracy of NCD-based classification methods is not predictable from compression rate alone when using neural compressors in the text classification domain. Second, by comparing different LLMs as neural compressors, we show that models of similar size but different architectures and pretraining have similar 
performance when compared to traditional compressors. Third, the extent to which Neural NCD improves or hinders performance over a Euclidean distance approach using the model's latent representations is model-dependent; we show that Neural NCD performs better than latent representations for certain models and performs worse for others.

We investigate Neural NCD for non-parametric NCD-based text classification in a few-shot setting using the datasets AGNews, 20News, and DBpedia (though we only use AGNews and DBpedia in \autoref{other_models} and \autoref{latent}). We use Bellard's \textit{ts\_zip} utility \cite{bellard_ts_zip} to perform neural compression with the model RWKV 169M \cite{peng2023rwkvreinventingrnnstransformer}. We compare these results with the use of the traditional compressors \textit{gzip}, \textit{zstandard} (\textit{zstd}), and \textit{lzma} on the same task. 

To investigate the few-shot setting, we draw $n=\{5,10,50,100\}$ labeled train samples from each class, obtain a subset of 100 test samples via stratified sampling across five trials for each $n$, calculating the mean test accuracy, 95\% confidence interval, and compression rate (compressed size / original size). The $O(n^2)$ computational complexity of pairwise NCD calculation for neural compression is a limiting factor in the computational cost of this method, scaling with LLM model size and size of training and test sets.  For this reason we subsample only 100 test samples here for each of the five trials. See \autoref{appendix_1} for further experiment details.

\subsection{Neural compressor has variable Neural NCD performance despite consistently superior compression rates.}

It has been hypothesized that compressors that achieve better compression rates should improve the performance of NCD-based methods through a better approximation of Kolmogorov Complexity \cite{jiang-etal-2023-low, jiang_few-shot_2022}. Through our text classification experiments with a neural compressor, we find that compression rate alone is not enough to predict the differences in NCD-based classification accuracy when using neural compressors across various datasets in the text domain. We note the compression rates (lower is better compression) for each compressor on each dataset in \autoref{tab:compression_rates}. 

\begin{wraptable}{R}{0.5\textwidth}
    \caption{Compression rates (lower is better compression) across compressors for the AGNews, 20News, and DBpedia datasets. For neural compressors, we use the raw compression rate which ignores the size of the neural network. The RWKV 169M model achieves the best compression rate across each dataset.}
\begin{tabular}{@{}lccc@{}}
\toprule
\multicolumn{1}{c}{\textbf{Compressor}} & \textbf{AGNews} & \textbf{20News} & \textbf{DBpedia} \\ \midrule
gzip                                    & 0.785           & 0.593           & 0.824            \\
zstd                                    & 0.743           & 0.596           & 0.785            \\
lzma                                    & 1.070           & 0.685           & 1.115            \\
RWKV 169M                               & \textbf{0.248}  & \textbf{0.204}  & \textbf{0.248}   \\ \bottomrule
\end{tabular}%
    \label{tab:compression_rates}
\end{wraptable}

\begin{figure}[!ht]
    \centering
    \begin{subfigure}[b]{0.32\textwidth}
        \centering
        \caption{AGNews}

        \includegraphics[width=\textwidth]{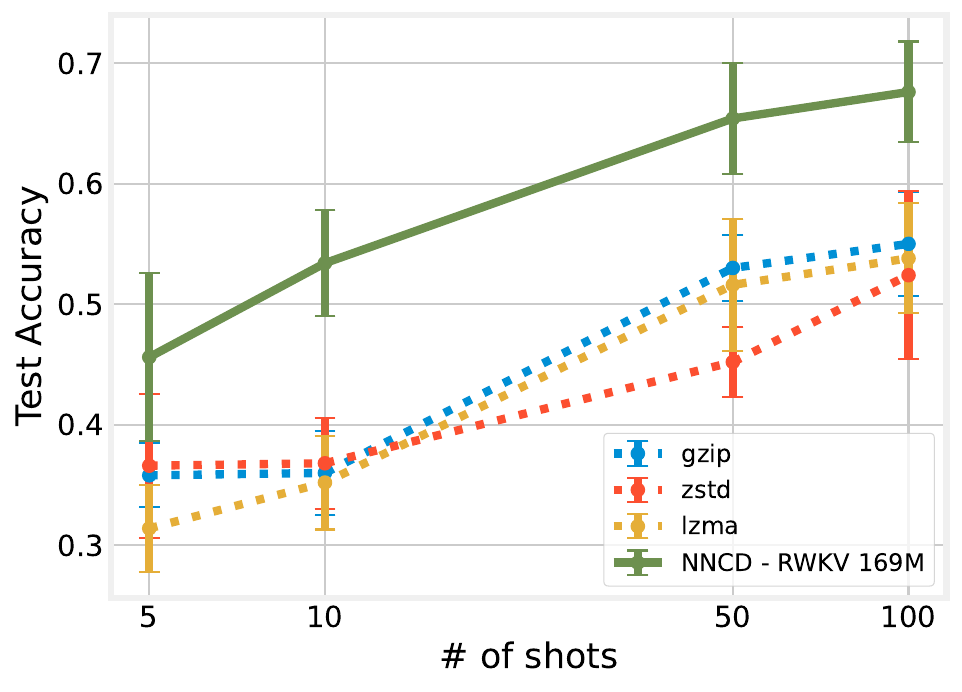}
    \end{subfigure}
    \hfill
    \begin{subfigure}[b]{0.32\textwidth}
        \caption{20News}
        \centering
        \includegraphics[width=\textwidth]{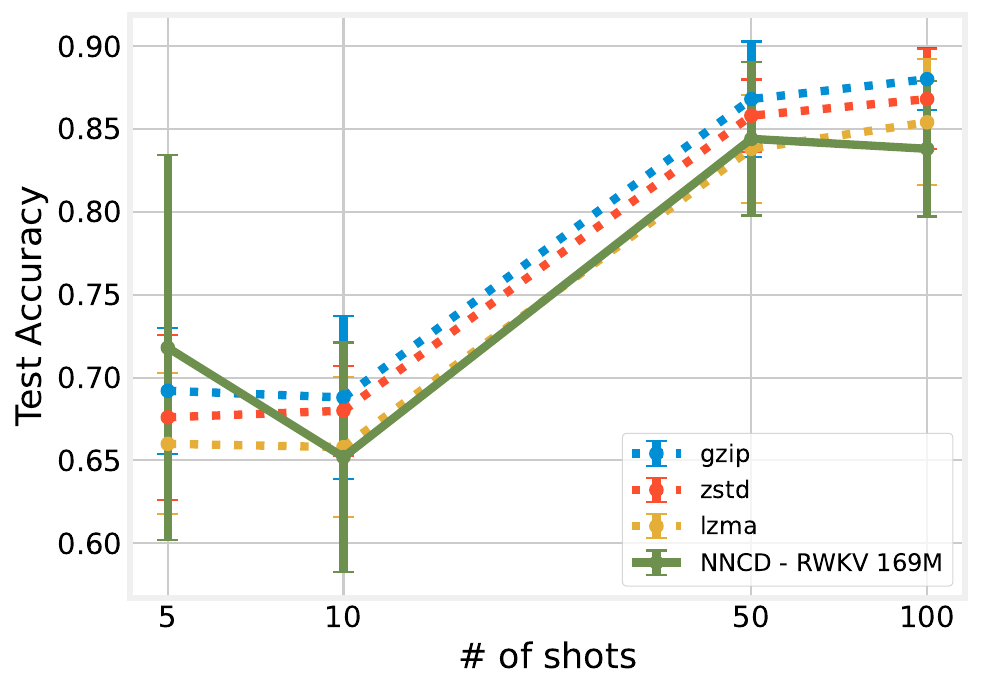}
    \end{subfigure}
    \hfill
    \begin{subfigure}[b]{0.32\textwidth}
        \centering
        \caption{DBpedia}
        \includegraphics[width=\textwidth]{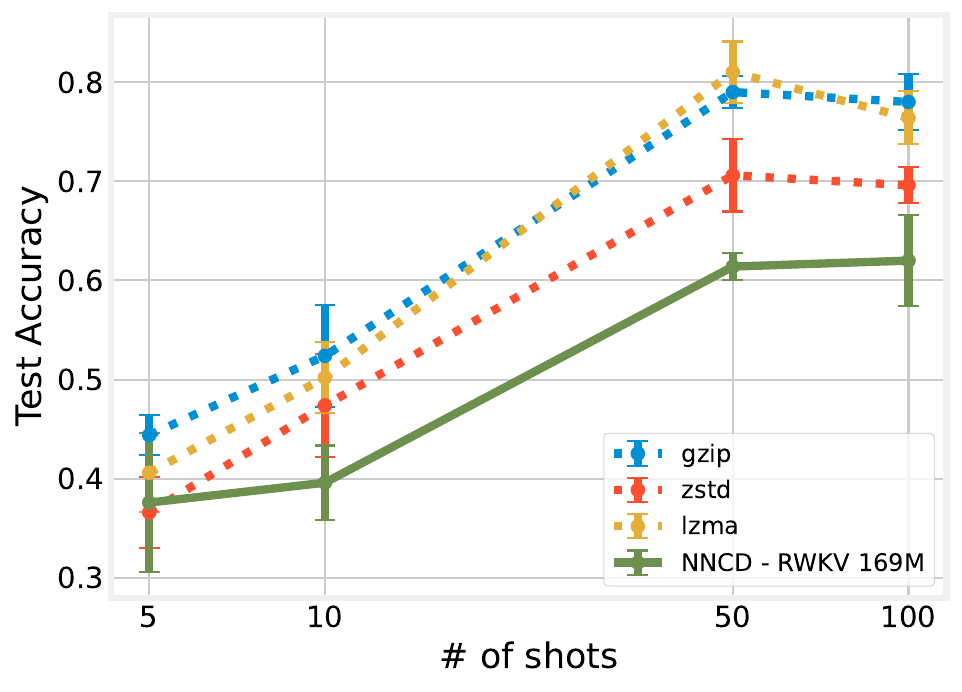}
    \end{subfigure}
    
    \caption{Comparison of RWKV 169M neural compressor and traditional compressors using $k$NN with NCD across the datasets AGNews, 20News, and DBpedia (NNCD = Neural NCD). Despite the neural compressor achieving superior compression rates, we find cases where Neural NCD outperforms, underperforms, and performs on-par with traditional compressors on the few shot sequence classification task. This calls into question the hypothesis that accuracy of NCD-based methods is predictable solely from compression rates.}
    \label{fig:neural_vs_trad}
\end{figure}

Ignoring the size of the language models, even a small, simple model such as RWKV 169M achieves significantly lower compression rates than traditional compressors. Despite the RWKV neural compressor achieving superior compression rates than traditional compressors on each dataset, Neural NCD yields varying relative performance when compared to using NCDs of traditional compressors, outperforming them on AGNews, performing on-par with them on 20News, and underperforming them on DBpedia (see \autoref{fig:neural_vs_trad}). Of particular note is that the RWKV compressor and \textit{gzip} retain their compression rates across both AGNews and DBpedia, and yet neural compression outperforms \textit{gzip} on the former and underperforms it on the latter. 

\begin{figure}[!h]
    \centering
    \includegraphics[width=0.75\linewidth]{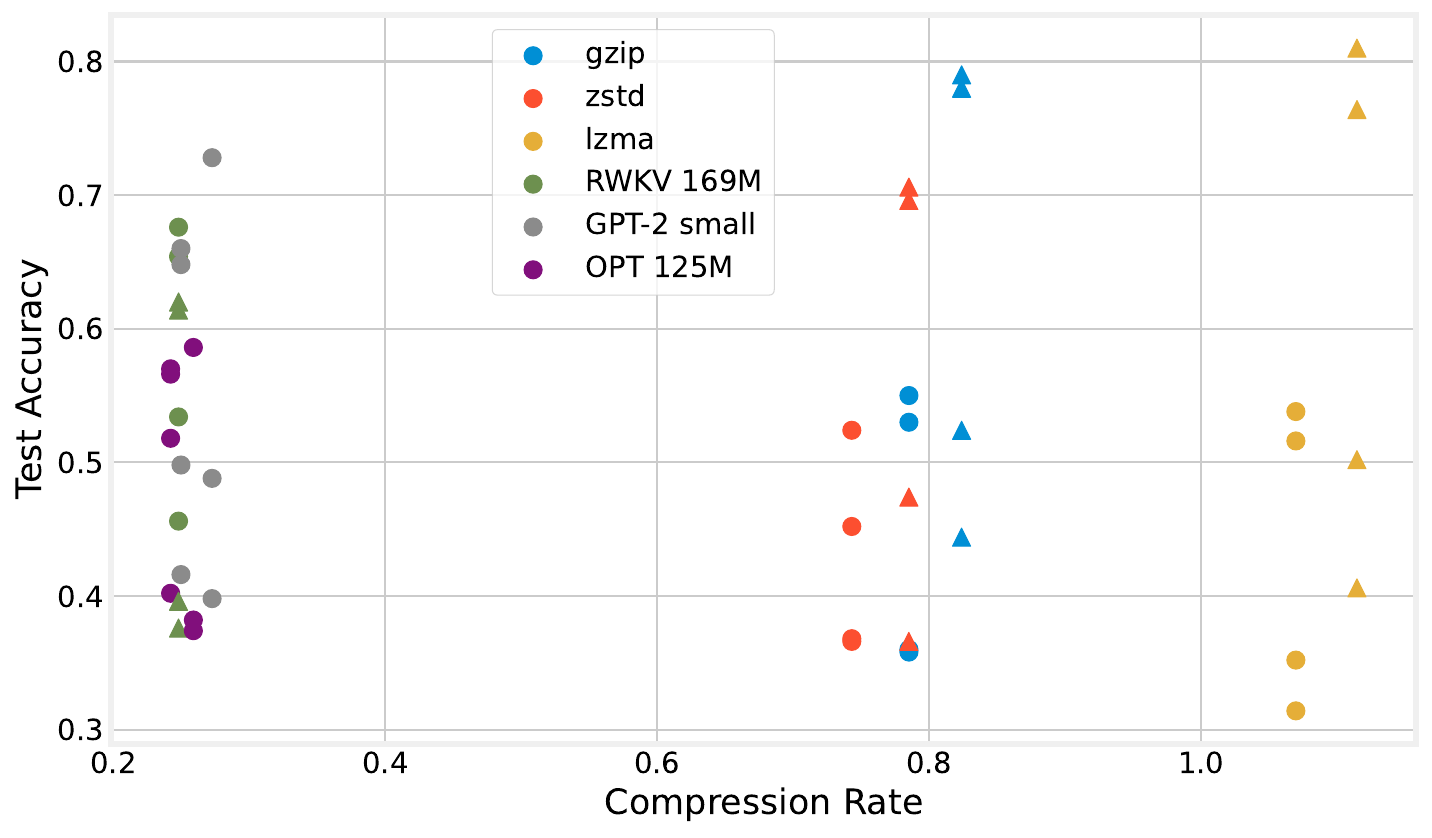}
    \caption{Test accuracy plotted against compression rate (lower is better compression) for AGNews and DBpedia across different few shot settings. Different shapes indicate different datasets, and each compressor is its own color. If compression rate and predictive performance were correlated, we would expect a diagonal relationship to occur, but none exists. }
    \label{fig:comp_rate_vs_acc}
\end{figure}

We see the same phenomenon strictly among the traditional compressors as well, where there is no big difference in performance despite varying compression rates \footnote{Note that for short sequences, LZMA's dictionary size outweighs the compression savings, leading to rates greater than 1. However, despite these poor compression rates LZMA is still able to do NCD-based sequence classification on-par with \textit{gzip} and \textit{zstd}.}. We plot test accuracies across compression rates for AGNews and DBpedia across all few-shot settings in \autoref{fig:comp_rate_vs_acc}. We can see that each compressor tends to have its own range of compression rates for these datasets, and despite drastically varying rates, the test accuracy spread is roughly the same, even with neural compressors with much lower compression rates.

The internal workings of each compression algorithm are important in determining NCD quality. For example, \textit{gzip} searches for repeated byte sequences only within a 32 KB sliding window. Since the concatenation of two inputs is a crucial component to calculating the NCD, large sequence lengths will prevent \textit{gzip} from exploiting far-reaching redundancies between the two. Differences in neural network architectures among neural compressors may yield different abilities to identify these long range similarities and thus varying NCD quality.

\begin{wraptable}[16]{R}{0.4\textwidth}  
\vspace{-15pt} %
\centering
\caption{
Compression rates when using various pretrained language models for neural compression on the AGNews and DBpedia datasets. For neural compressors, we use the raw compression rate which ignores the size of the neural network.}
\label{tab:other_models_compression_rates}
\begin{tabular}{@{}lcc@{}}

\toprule
\multicolumn{1}{c}{\textbf{Compressor}} & \textbf{AGNews}  & \textbf{DBpedia} \\ \midrule
gzip                                    & 0.785                            & 0.824            \\
RWKV 169M                               & 0.248                            & 0.248            \\
GPT-2 117M                              & 0.250                            & 0.273            \\
OPT 125M                                & 0.242                            & 0.259 \\ \bottomrule
\end{tabular}%
\end{wraptable}

\subsection{Using other models for neural compression} \label{other_models}

\begin{figure}[!ht]
    \centering
    \begin{subfigure}[b]{0.45\textwidth}
        \centering
        \caption{AGNews}
        \includegraphics[width=\textwidth]{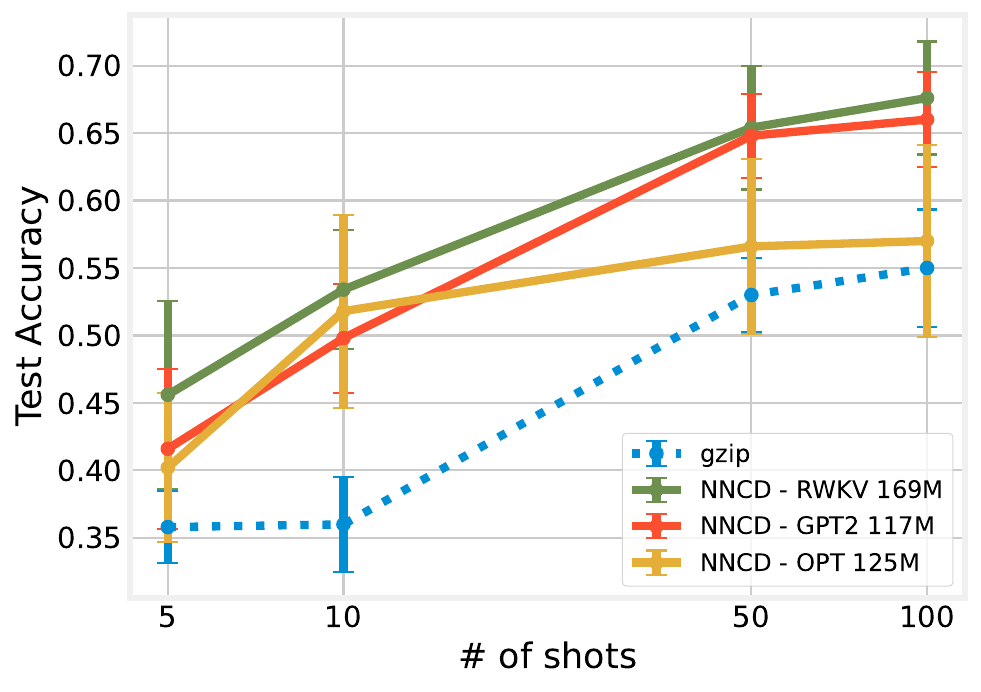}
    \end{subfigure}
    \hfill
    \begin{subfigure}[b]{0.45\textwidth}
        \caption{DBpedia}
        \centering
        \includegraphics[width=\textwidth]{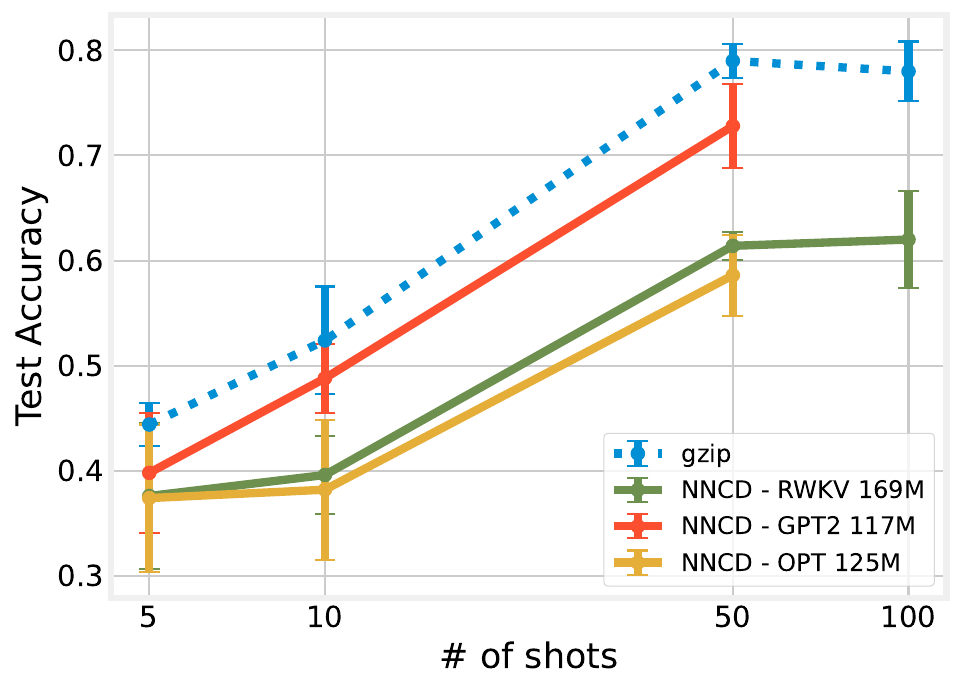}
    \end{subfigure}
    \hfill
    
    \caption{Comparison of RWKV 169M, GPT-2 117M, and OPT 125M, as the neural compressors used for Neural NCD. Neural compressors outperform \textit{gzip} similarly on AGNews, and underperform it similarly on DBpedia. }
    \label{fig:other_models}
\end{figure}

In order to investigate the effects of the particular choice of model and model architecture on Neural NCD performance, we run identical experiments as above, swapping out RWKV 169M for two other models: GPT-2 117M (GPT-2 small) \cite{radford2019language} and OPT 125M \cite{zhang2022optopenpretrainedtransformer}, as shown in \autoref{fig:other_models}. Here we only use the \textit{gzip} results out of the traditional compressors since they perform comparably and to aid visual clarity. We can see that all neural models outperform \textit{gzip} in the same way on AGNews, and underperform it in the same way on DBpedia. Compression rates across models are shown in \autoref{tab:other_models_compression_rates}.

\subsection{Comparison with Euclidean distance between Latent Representations} \label{latent}

\begin{figure}[!h]
    \centering
    \begin{subfigure}[b]{0.45\textwidth}
        \centering
        \caption{AGNews}
        \includegraphics[width=\textwidth]{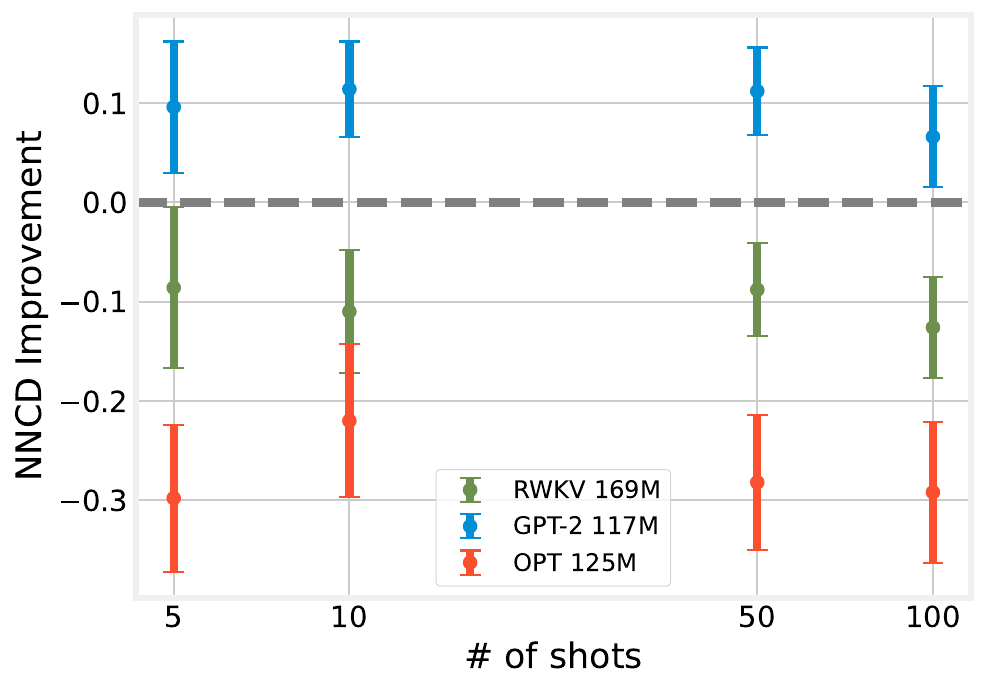}
    \end{subfigure}
    \hfill
    \begin{subfigure}[b]{0.45\textwidth}
        \caption{DBpedia}
        \centering
        \includegraphics[width=\textwidth]{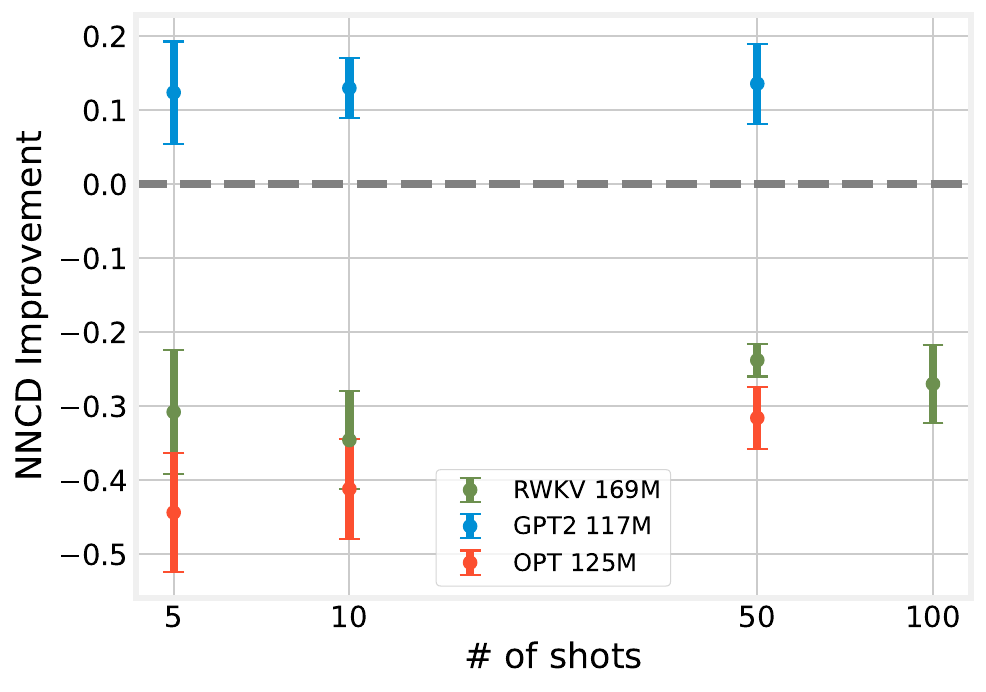}
    \end{subfigure}
    \hfill
    
    \caption{Test accuracy difference when comparing Neural NCD to Euclidean distance on sequence latent representations with 95\% confidence interval. Values above 0 indicate Neural NCD outperforming Euclidean distance. For RWKV, the representation is the final hidden state. For GPT2 and OPT, we average the latent representation of each token. Despite comparable compression rates of each model, the quality and usefulness of distance between latent representations is highly variable across models. 
    }
    \label{fig:latent_representations}
\end{figure}

For each neural compressor we test, we perform identical experiments as previously mentioned using the model's latent representation of a sequence and $k$NN with Euclidean distance. The results are shown in \autoref{fig:latent_representations}. We find that depending on the choice of neural network model used for neural compression, Neural NCD can either outperform or underperform the Euclidean distance-based approach on sequence latent representations. For example, a Neural NCD approach with GPT-2 small outperforms using the model's latent representations, however with OPT 125M, latent representations drastically outperform a Neural NCD approach. This indicates that despite various LLMs providing similar compression ability, the usefulness of their latent representations for Euclidean distance-based approaches can differ.

\section{Conclusion} \label{conclusion}
We have shown that compression rate alone does not reliably predict NCD-based classification accuracy when using neural compressors in the text domain.  With a variety of neural network architectures with which the machine learning community trains language models and neural compressors, it is likely that these architectural differences will also yield varying ability to exploit compressible redundancies across concatenated sequences yielding varying NCD quality. We've shown that neural models with different architectures and pretraining details tend to overperform or underperform traditional compressors in similar ways on the same dataset. Finally, we compared Neural NCD with latent representations from the same model, showing that Neural NCD outperforms latent representations for some models and under-performs them for others.

\bibliographystyle{ACM-Reference-Format}
\bibliography{references}

\clearpage
\appendix
\section{Experiment Details} \label{appendix_1}
We use the labeled datasets AGNews, 20News, and DBpedia. For AGNews and DBpedia, we use the original dataset and classification setting: four classes for AGNews and fourteen classes for DBpedia. For 20News, we reduce the number of classes to two using only \textit{alt.atheism} and \textit{comp.graphics}. For $k$NN classification algorithm, we use $k=3$ for all experiments.

\end{document}